# MaScQA: A Question Answering Dataset for Investigating Materials Science Knowledge of Large Language Models

Mohd Zaki[1], Jayadeva[2], Mausam[3,4], N. M. Anoop Krishnan[1,3]

[1]Department of Civil Engineering, Indian Institute of Technology Delhi, Hauz Khas, New Delhi 110016, India

[2]Department of Electrical Engineering, Indian Institute of Technology Delhi, Hauz Khas, New Delhi 110016, India

[3]Yardi School of Artificial Intelligence, Indian Institute of Technology Delhi, Hauz Khas, New Delhi 110016, India

[4]Department of Computer Science & Engineering, Indian Institute of Technology Delhi, Hauz Khas, New Delhi 110016, India

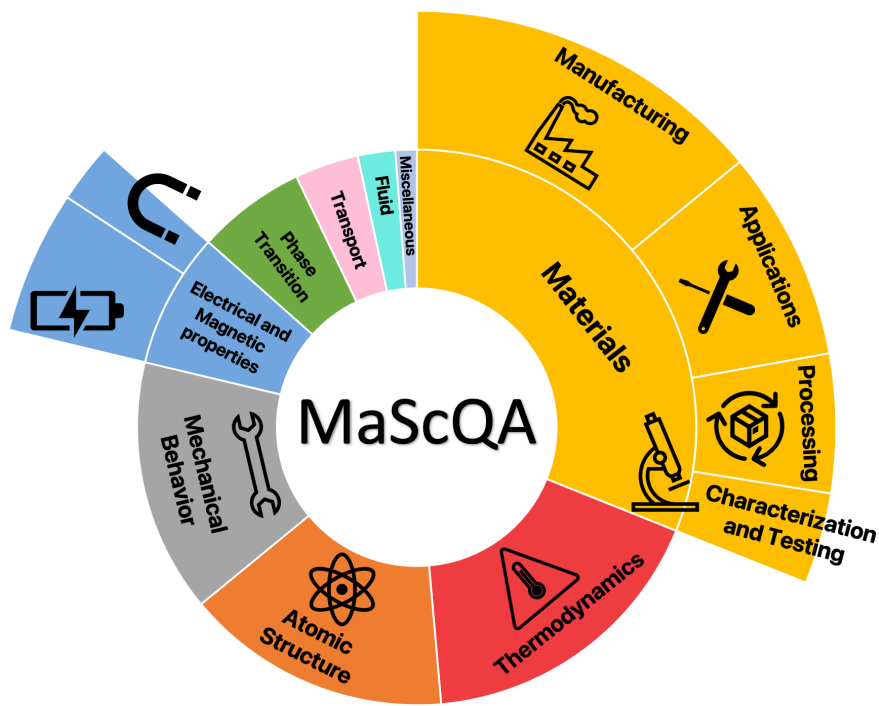

**Abstract**

Information extraction and textual comprehension from materials literature are vital for developing an exhaustive knowledge base that enables accelerated materials discovery. Language models have demonstrated their capability to answer domain-specific questions and retrieve information from knowledge bases. However, there are no benchmark datasets in the materials domain that can evaluate the understanding of the key concepts by these language models. In this work, we curate a dataset of 650 challenging questions from the materials domain that require the knowledge and skills of a materials student who has cleared their undergraduate degree. We classify these questions based on their structure and the materials science domain-based subcategories. Further, we evaluate the performance of GPT-3.5 and GPT-4 models on solving these questions via zero-shot and chain of thought prompting. It is observed that GPT-4 gives the best performance (~62% accuracy) as compared to GPT-3.5. Interestingly, in contrast to the general observation, no significant improvement in accuracy is observed with the chain of thought prompting. To evaluate the limitations, we performed an error analysis, which revealed conceptual errors (~64%) as the major contributor compared to computational errors (~36%) towards the reduced performance of LLMs. We hope that the dataset and analysis performed in this work will promote further research in developing better materials science domain-specific LLMs and strategies for information extraction.

***Keywords. Large language models, materials science, materials discovery, the chain of thought***

**Introduction**

Large language models (LLMs) are machine learning (ML) models based on transformer neural network architecture [1]. These models are called *large* due to their billions of inherent parameters. The increase in the number of model parameters and different training strategies have improved the performance of these models on natural language tasks such as question answering[2,3], text summarization [4,5], sentiment analysis[1,3], machine translation[6], conversational abilities[7–9], and code generation[10]. Numerous datasets allow researchers to benchmark the performance and evaluate the different capabilities of LLMs. MMLU, a question-answering dataset, comprises questions under four broad categories of humanities, social sciences, STEM, and others. The categories have questions from domains like high school subjects, clinical knowledge, and mathematics, to name a few[11,12]. Another dataset is HellaSwag[13], a benchmark dataset for common sense natural language inference, where the input is a sentence, and the output should be a sentence that completes the given input. A similar dataset, WinoGrande[14], was proposed by Sakaguchi et al. (2020), having 273 expert-crafted pronoun resolution questions. HumanEval[10] is a dataset created to evaluate the performance of LLMs in writing codes. Dua et al. (2019) proposed the DROP[15] dataset for assessing the performance of LLMs on reading comprehension tasks. To evaluate the performance of LLMs on grade school mathematics problems, Cobbe et al. (2021) introduced the GSM8K[16] dataset, comprising linguistically diverse mathematical word problems. AI2 Reasoning Challenge (ARC) contains school-level science-based questions which have been used as a dataset to show the state-of-the-art performance achieved by GPT-4[17]. However, based on literature review and analysis of the technical report on GPT4 and research papers that introduced other LLMs like Chinchilla[18] and PaLM[2] reveals that there are no datasets related to materials science on which these LLMs have been benchmarked yet.

The datasets which exist in the materials science domain are mainly for tasks like named entity recognition (NER)[19,20], classification[21–23], synthesis process and relation classification[24], composition extraction from tables[25], which researchers use to benchmark the performance of materials domain LLMs. These models, namely, MatSciBERT[22] (first materials-domain language model), MatBERT[26], MaterialsBERT[27], OpticalBERT[28], and BatteryBERT[23] have been trained on domain-specific texts, which resulted in state-of-the-art results on the tasks mentioned above. However, there are no large and diverse datasets in the materials domain that can be used for evaluating the natural language question-answering ability of LLMs. The development of such a dataset is thus crucial to investigate the materials domain knowledge of these LLMs so that they can be further used for addressing challenging problems related to materials discovery for areas such as manufacturing, energy, environment, and sustainability. This information is further important to understand the lacunae of the understanding of such LLMs, which are being proposed to be used for several domains such as manufacturing, planning, material synthesis, and materials discovery[22,27].

To address this challenge, we present a question-answering dataset on the materials domain. Specifically, we try to answer the following questions in this paper:

1. How well do general-purpose LLMs perform in answering complex questions from the materials science domain?
2. Can we improve the performance of the LLMs by using the chain of thought prompting methods?
3. What are the factors limiting the performance of these LLMs on this dataset?

To this end, we collected questions that require students to have a graduate-level understanding of material science topics to solve them. These questions and answers are carefully curated from the original questions in the graduate aptitude test in engineering (GATE) exam—a national-level examination for graduate admission in India. More than 800,000 students give this exam annually, with an average of 100,000 students in major disciplines, such as mechanical or civil engineering, to enroll in masters/doctoral courses in the premier institutes in India. We classify these questions based on their (a) structure and (b) domain knowledge required to solve them. We then evaluate the performance of state-of-the-art proprietary models: GPT-3.5 and GPT4, in solving these questions. We used the API of these models to obtain answers to the questions in two ways: first, by directly prompting the models to answer the questions (zero-shot prompting), and second, by asking the models to solve the questions step by step, also known as the *Chain of Thought* prompting[29]. The availability of MaScQA will allow the researchers to benchmark existing models and prompting strategies. Specifically, the analysis from a domain-specific perspective will allow the researchers to train better domain-specific LLMs and help them decide where these models can be used in the materials discovery pipeline.

**Methodology**

*Dataset preparation*

We are motivated to investigate how LLMs will perform on questions that require an undergraduate-level understanding of materials science topics for their solution. To compile a dataset of such questions, we take question papers related to materials science and metallurgical engineering asked in the GATE examination conducted in India for admission to masters and doctorate courses. To this end, we compiled 650 questions and classified them into four types based on their structure: Multiple choice questions (MCQs), Match the following type questions (MATCH), Numerical questions where options are given (MCQN), and numerical questions (NUM). MCQs are generally conceptual, given four options, out of which mostly one is correct and sometimes more than one option is also correct (Fig. 1 (a)). In MATCH, two lists of entities are given, which are to be matched with each other. These questions are also provided with four options, out of which one has the correct set of matched entities (Fig. 1 (b)). In MCQN, the question has four choices, out of which the correct one is identified after solving the numerical stated in the question (Fig. 1 (c)). The NUM type questions have numerical answers, rounded off to the nearest integer or floating-point number as specified in the questions (Fig. 1 (d)).

To understand the performance of LLMs from a domain perspective, we classified the questions into 14 categories. The list of categories was prepared in consultation with domain experts who teach materials science subjects at the institute where this research is conducted. Then all the questions are assigned one of the categories by two experts. The conflict in the category assignments was resolved through discussion and mutual agreement. Figure 2 shows the number of questions in each category. The color of the bars represents the broad category of materials science topics under which each subtopic is shown in the graphical abstract. The database can be accessed at https://github.com/M3RG-IITD/MaScQA.

| Floatation beneficiation is based on the principle of<br>(A) Mineral surface hydrophobicity (B) Gravity difference<br>(C) Chemical reactivity (D) Particle size difference | A peak in the X-ray diffraction pattern is observed at $2\theta = 78°$, corresponding to {311} planes of an fcc metal, when the incident beam has a wavelength of 0.154 nm. The lattice parameter of the metal is approximately<br><br>(A) 0.6 nm (B) 0.4 nm (C) 0.3 nm (D) 0.2 nm |
|---|---|
| (a) Multiple choice question (**MCQ**) | (c) Numerical question with multiple choices (**MCQN**) |
| Match the processes given in Group I with the corresponding metals in Group II:<br>Group I: [P. Matte smelting, Q. Cyanide leaching, R. Carbothermic reduction, S. Fused salt electrolysis]<br><br>Group II: [1.Lead, 2. Copper, 3. Gold, 4. Aluminium]<br>Options:<br>(A) P-1, Q-2, R-1, S-4 (B) P-2, Q-3, R-1, S-4<br>(C) P-2, Q-1, R-3, S-4 (D) P-2, Q-3, R-4, S-1 | The third peak in the X-ray diffraction pattern of a face-centered cubic crystal is at $2\theta$ value of 45°, where $2\theta$ is the angle between the incident and reflected rays. The wavelength of the monochromatic X-ray beam is 1.54 Å. Considering first-order reflection, the lattice parameter(in Å) of the crystal is? (Round off to two decimal places) |
| (b) Matching type question (**MATCH**) | (d) Numerical question (**NUM**) |

**Figure 1.** Sample questions from each category (a) multiple choice question (MCQ), (b) matching type question (MATCH), (c) numerical question with multiple choices (MCQN), and (d) numerical question (NUM).

*Solutions using LLMs*

In this work, we benchmark the question-answering ability of GPT-3.5 and GPT-4 models on the MaScQA dataset. The questions are provided to each model in two ways: first, directly asking the model to solve the question, and second, asking the models to solve the given question by providing a detailed step-by-step solution. We call the first approach zero-shot question answering and the second approach chain of thought (CoT) reasoning[29]. The questions are fed to the model using the OpenAI API and selecting the appropriate model type. The prompt used in the first approach is *"Solve the following question. Write the correct answer inside a list at the end."* for the second approach, the prompt is *"Solve the following question with highly detailed step-by-step explanation. Write the correct answer inside a list at the end."* The last sentence in the prompt was used to automatically retrieve the correct option/answer from the model output and match it with the answer key. However, the model did not always give output in the desired format. Hence, the entire model output is saved as a text file which was then used for manually extracting the answers for comparing with the actual answers provided in the official answer keys of the respective papers. The solutions to all the questions obtained using two approaches for both models can be accessed at https://github.com/M3RG-IITD/MaScQA. The official answer keys are obtained from the official website of IIT Kharagpur, which is one of the organizing institutes of the GATE exam. https://gate.iitkgp.ac.in/old_question_papers.html. The LLMs' performance on two prompting methods is discussed in detail in the following sections.

**Results**

Figure 2 shows the details of the dataset comprising a total of 650 questions in different categories. First, we categorize the questions based on their structure. We observe that largest the category of questions (287) are MCQs, while 70 are MATCH-type questions. Further, 66 questions are MCQN, while the remaining 229 questions are NUM that do not provide any choices. Further, we analyze different materials domains covered by this set of questions. To this extent, the questions are categorized into 14 domains: thermodynamics, atomic structure, mechanical behavior, materials manufacturing, material applications, phase transition, electrical properties, material processing, transport phenomenon, magnetic properties, material characterization, fluid mechanics, material testing, and miscellaneous.

Figure 2 shows the number of questions in different domain-specific categories. To visualize the frequently used words related to each domain-specific category of questions, word clouds are shown in Figure 3. The maximum number of questions (114) are in the *thermodynamics* category, which deals with questions related to enthalpy of formation, energy balance during chemical reactions, transition temperatures, activation energy, and heat transfer (Fig. 3(a)). The category of *atomic structure* comprises 100 questions which are based on concepts such as dislocations, diffraction planes, and crystal structures (Fig. 3(b)). The *mechanical behavior* category is based on the concepts of stress-strain behavior of materials, creep, fatigue, and fracture mechanics (Fig. 3(c)). In *materials manufacturing* (Fig. 3(d)) and *material applications* (Fig. 3(e)), the questions test the knowledge of extraction processes of materials from their respective ores and why a particular material is used for a specific application. Thus, these questions require logical understanding connecting multiple concepts: first, "recall" or "deduce" the properties of a material based on its composition, label, or processing conditions, and second, "identify" the properties required for a particular application and then connect these two concepts to "derive" a logical explanation to arrive at the correct answer. The questions on *phase transition* test the knowledge of how phase transition can be induced in materials, how to calculate the percentage of different phases in the materials, and the characteristics of different phases. This is also indicated by the high frequency of words related to different phases of materials (Fig. 3(f)). The questions on *electrical properties* include fuel cells, characteristics of materials used in batteries, and semiconductor devices (Fig. 3(g)). Then, questions are based on *material processing* such as annealing, tempering, recrystallization, welding, etc. (Fig. 3(h)). The questions on transport phenomena test concepts related to the diffusion or transport of ions (Fig. 3(i)). The question related to *magnetic properties* tests the knowledge about magnetization and the characteristics of different magnetic materials (Fig. 3(j)). The material characterization topic has questions related to methods like scanning electron microscopy, diffraction studies, and back scattered electron microscopy (Fig. 3(k)). The fluid mechanics topic comprises questions on the viscosity of the fluid and the movement of particles in a viscous medium (Fig. 3(l)). In the *material testing* topic, the questions are based on mostly non-destructive material testing methods (Fig. 3(m)). The miscellaneous category deals with questions requiring simultaneously understanding multiple materials science domains for their solution (Fig. 3(n)).

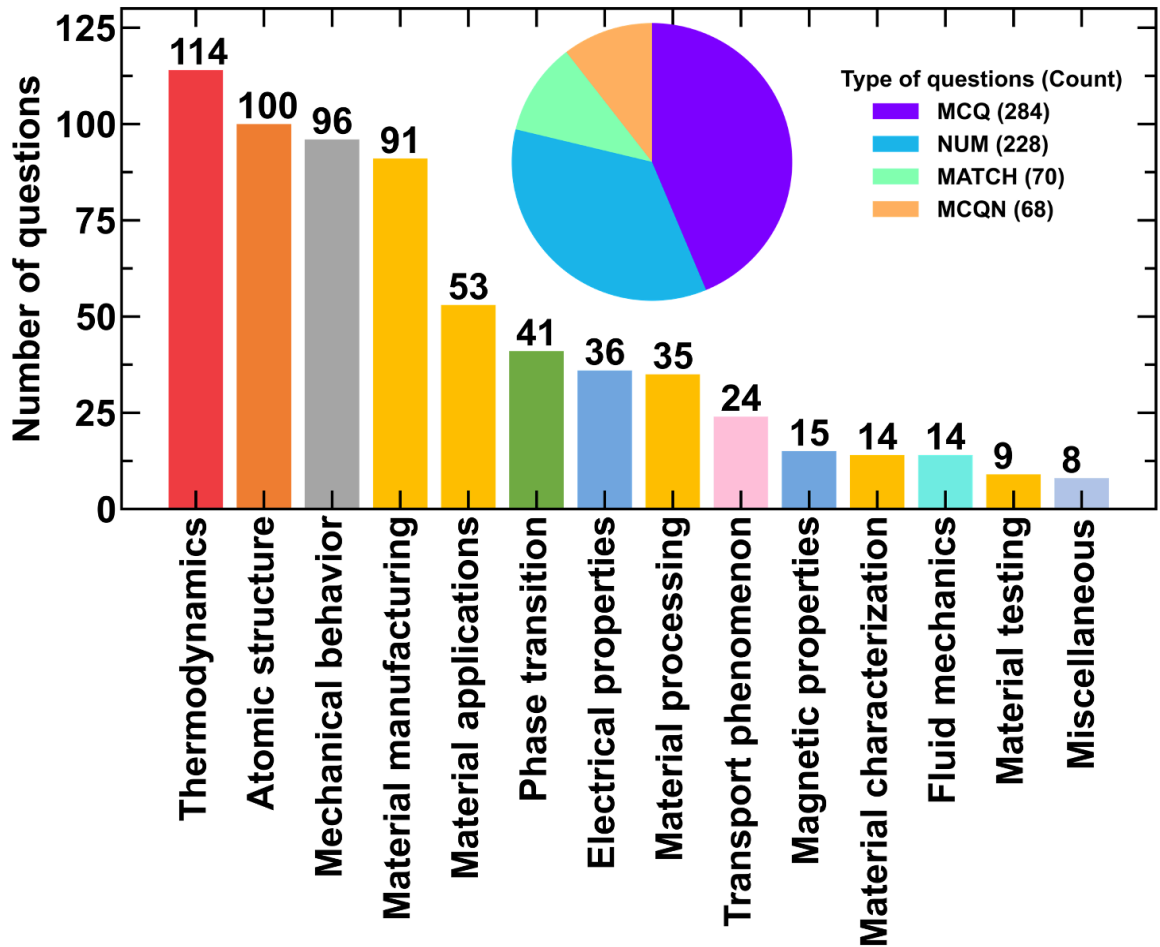

**Figure 2**. The number of questions in each materials science sub-domain. The bar chart shows the distribution of questions in different sub-domains. The pie chart shows the number of questions classified according to question structure.

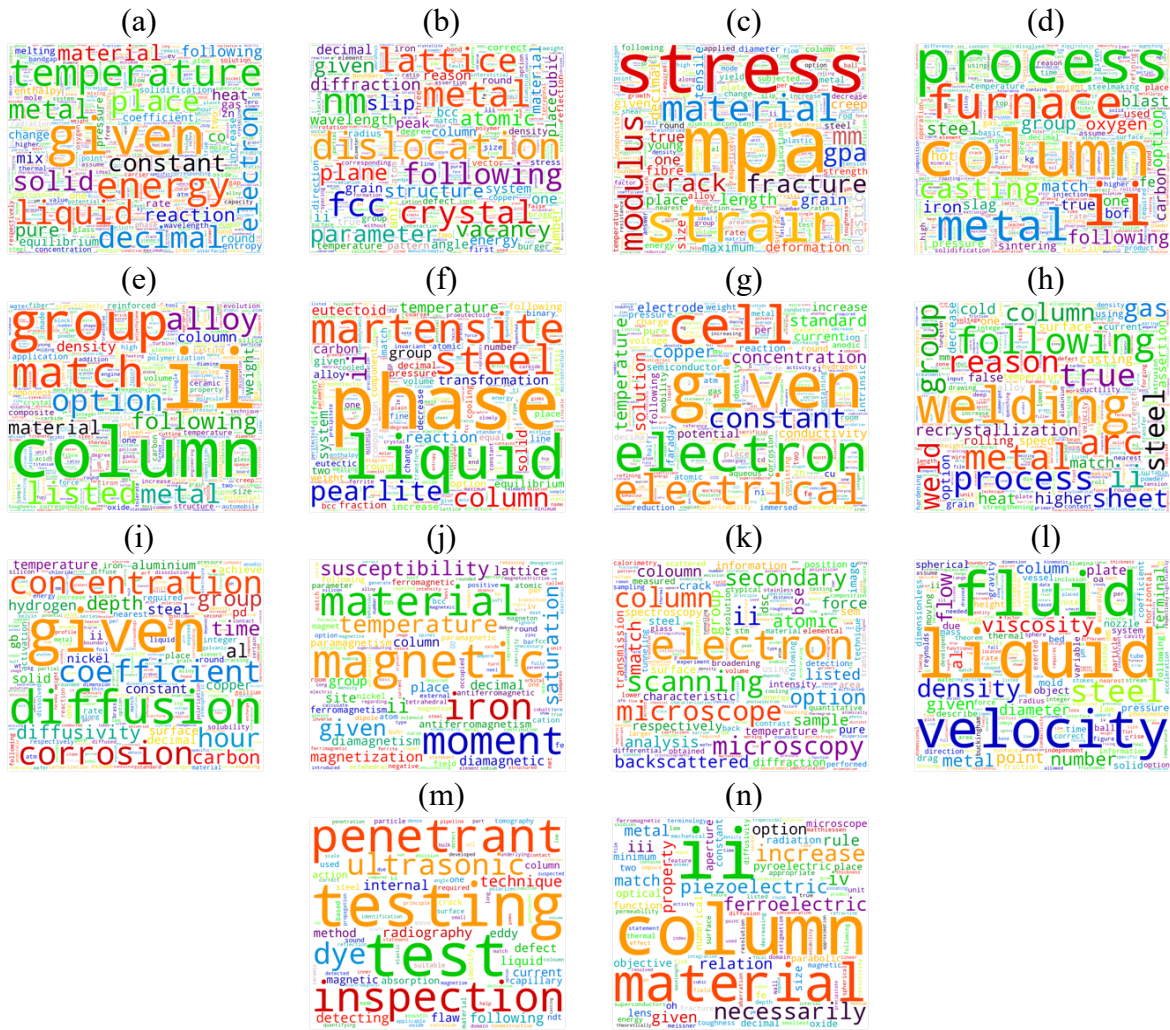

**Figure 3.** Word-cloud for different topics in MaScQA (a)Thermodynamics, (b)Atomic structure, (c)Mechanical behavior, (d) Material manufacturing, (e) Material applications, (f) Phase transition, (g) Electrical properties, (h) Material processing, (i) Transport phenomena (j) Magnetic properties (k) Material characterization (l) Fluid mechanics (m) Material testing (n) Miscellaneous. The size of the word in a cloud is proportional to the occurrence of the word in all the questions for a given topic.

Now, we evaluate the performance of LLMs on MaScQA and the effect of prompting methods on the performance, corresponding to the first two questions posed in this work. Table 1 reports the accuracy of the LLMs on the MaScQA corpus. The scores corresponding to model names GPT-3.5 and GPT-4 represent the accuracy of the models when questions are asked directly to the models representing zero-shot answering. The model names with the suffix "CoT" implies we have asked the models to provide detailed "stepwise" solutions to the given questions. In MCQs, we observe that GPT-4 significantly outperforms GPT-3.5. Further, we also observe that the CoT provides only marginal improvement in the result for GPT-3.5 and GPT-4. Here, GPT-4-CoT gives an accuracy of 77.11%, which is a high score considering the difficulty levels of this exam. Also, the performance of GPT-4-CoT is ~20% higher than GPT-3-CoT for MCQ type of questions. For MATCH questions, GPT-4-CoT exhibits the maximum performance with a score of 92.86%, a very high score considering the amount of knowledge required to connect the entities. In contrast, the variants of GPT-3.5 performed poorly on MATCH questions, with a score of 40% and 38.57% for the variants without and with CoT, respectively. In this case, the GPT-4-CoT provides ~4% improvement over direct prompting. For MCQN, GPT-4 gives the best performance with a score of 58.82%, while CoT reduces the model's performance to 51.47%. The same trend of reduced performance on these questions is observed with the GPT-3.5 model. This implies that CoT prompting may not always lead to better performance. Now, we focus on the numerical questions. Among all the categories, models exhibit the worst performance in the NUM category. Here, GPT-4 and GPT-4-CoT obtain the maximum score of 33.33% and 36.84%. Interestingly, we observe that CoT yields poorer results in the case of GPT-3.5, while it yields better accuracy in the case of GPT-4. Finally, regarding overall performance, GPT-4-COT gives the best score of 62%, with GPT-4 following closely at 60.15%. It should be noted that in MCQ, there are 13 questions where more than one options are correct, of which GPT-4 and GPT-4-CoT answered six and seven questions correctly, respectively. Interestingly, we observe that CoT does not always give improved results. In fact, for GPT-3.5, CoT gives poorer results in all the cases except MCQs and marginally better results for GPT-4 in all the cases except MCQN. Note that this observation contrasts with the general observation that the CoT prompting results in improved performance of LLMs on QA tasks. This is further evaluated in detail later.

| Evaluation Method | MCQ (285) | Matching (MATCH) (70) | Numerical with MCQ (MCQN) (67) | Numerical (NUM) (228) | Overall accuracy |
|---|---|---|---|---|---|
| Baseline scores | 25 | 25 | 25 | 0 | |
| GPT-3.5 | 56.49 | 40.00 | 35.82 | 15.79 | 38.31 |
| GPT-3.5-CoT | 56.84 | 38.57 | 34.33 | 14.04 | 37.38 |
| GPT-4 | 74.74 | 88.57 | **59.7** | 33.77 | 60.15 |
| GPT-4-CoT | **76.84** | **92.86** | 52.24 | **37.28** | **62.0** |

**Table 1.** Performance (% accuracy) of different evaluation styles using GPT models on various question types. The number in parenthesis represents the total number of questions under respective categories.

In addition to the performance of GPT models in answering different types of questions like multiple choice, numerical, and matching, which test different mental abilities of the students, it is also important to analyze the performance of the models from a domain perspective. To this end, we classify all the questions of our dataset into 14 broad categories. Table 2 shows the accuracy of the GPT-4-CoT prompting method while answering the questions.

It is observed that questions related to materials' mechanical and electrical behavior have the most percentage of incorrectly answered questions (~60%). The questions on thermodynamics, atomic structure, phase transition, transport phenomena, and magnetic properties have more than ~40% of incorrectly answered questions in the respective categories. Further, more than 15% of materials manufacturing, application, and characterization questions are incorrectly answered, with the lowest error rates for material characterization and no mistakes made on material testing questions. To further gain insights into the factors limiting LLMs' performance, we will discuss them by classifying the mistakes into two categories, as explained in the *Discussion* section.

| Category | Correct | | Incorrect | | Total |
|---|---|---|---|---|---|
| | # Questions | %age | # Questions | %age | |
| Thermodynamics | 63 | 55.26 | 51 | 44.74 | 114 |
| Atomic structure | 59 | 59.00 | 41 | 41.00 | 100 |
| Mechanical behavior | 43 | 44.79 | 53 | 55.21 | 96 |
| Material manufacturing | 62 | 68.13 | 29 | 31.87 | 91 |
| Material Applications | 46 | 86.79 | 7 | 13.21 | 53 |
| Phase transition | 25 | 60.98 | 16 | 39.02 | 41 |
| Electrical properties | 15 | 41.67 | 21 | 58.33 | 36 |
| Material processing | 31 | 88.57 | 4 | 11.43 | 35 |
| Transport phenomena | 15 | 62.50 | 9 | 37.50 | 24 |
| Magnetic properties | 9 | 60.00 | 6 | 40.00 | 15 |
| Material characterization | 10 | 71.43 | 4 | 28.57 | 14 |
| Fluid mechanics | 12 | 85.71 | 2 | 14.29 | 14 |
| Material testing | 9 | 100.00 | 0 | 0.00 | 9 |
| Miscellaneous | 5 | 62.50 | 3 | 37.50 | 8 |

**Table 2.** Performance of GPT-4-CoT on questions classified from materials science domain perspective

## Discussion

*Error Analysis*

To use LLMs effectively and to identify areas that require further research, it is important to understand the mistakes made by the LLMs in the materials domain. Answering a question requires retrieval of correct concepts/facts, applying them to the scenarios posed in the question by appropriate substitution in the relevant formulae, and then solving it correctly by applying relevant computational steps. To understand further, we can divide these errors into three categories, namely, (i) conceptual error: where the correct concept, equation, or facts related to the problem are not retrieved, or the LLM hallucinates some facts, (ii) grounding error: where the relevant concepts are not correctly applied to the scenario or incorrect values are substituted in the equations (for example, ºC to K conversion not applied) and (iii) computational error: where the numerical computation is performed incorrectly [32]. Note that CoT prompting enables the model to reflect upon the knowledge it already has, connect it with multiple choices, and then arrive at the answer. Thus, in general, it has been observed that CoT helps in reducing grounding errors (in our case, it virtually eliminates them).

To analyze different errors, we created a subset of 100 random questions where GPT-4-CoT answered incorrectly. Of these 100 questions, 54 are NUM, 27 are MCQs, 14 are MCQN, and five are matching-type questions (MATCH) (Table 3). All the questions From the domain-specific categories are included from those domains with less than ten mistakes made during GPT-4-CoT prompting (see Table 2). The remaining questions are randomly sampled from the other categories. The number of questions across materials science sub-domains in the subset of 100 questions is shown in Table 4. Note that there may be questions with conceptual and numerical errors, but we have considered only the conceptual error in these questions since it is the first to be found. If the retrieved concept is incorrect, we deem the computational error secondary.

Table 3 shows the errors made by GPT-4-CoT in different categories. The analysis of the 100 questions reveals that most errors are conceptual. Even in numerical problems, we observe that as many conceptual errors are made as numerical errors. It is interesting to observe that GPT-4-CoT is equally bad at retrieving concepts and doing calculations on NUM type questions. This explains the lowest performance of LLMs on these types of questions. Further, in MCQs and MATCH type questions, the error is always conceptual because answering such questions require retrieval of appropriate concepts and facts and then connecting them with relevant options. For MCQN, the computational error is more prevalent than the conceptual error. Most of the questions were answered incorrectly (64%) due to conceptual errors implying the need for domain-specific models or better prompting and problem-solving approaches.

As mentioned above, we observe that GPT-4-CoT makes no grounding errors. To evaluate whether this is due to the effectiveness of CoT, we investigate questions that are incorrectly answered by GPT-4 and correctly by GPT-4-CoT. Out of 65 questions from the entire dataset, GPT-4's solutions had ~70% conceptual errors, ~30% computational errors, and no grounding errors. Further, we also analyzed the errors made by GPT-4-CoT that are correctly answered by GPT-4. There were 53 such questions in the complete dataset. Out of these questions, solutions of 42 questions (~79%) had conceptual errors; for 1 question, there was a grounding error, and the remaining ten questions had computational errors when solved using GPT-4-CoT. Since there are little to no grounding errors in either GPT-4 or GPT4-CoT, both models are adept in this regard. The CoT prompting is helping reduce some numerical errors.

| Question Type | Conceptual error | | Computational error | |
|:---:|:---:|:---:|:---:|:---:|
| | # Questions | %age | # Questions | %age |
| MCQs | 27 | 100 | 0 | 0 |
| MATCH | 5 | 100 | 0 | 0 |
| MCQN | 5 | 35.71 | 9 | 64.29 |
| NUM | 27 | 50 | 27 | 50 |

**Table 3.** Types of the errors on 100 questions classified based on the structure

Table 4 shows the domain-wise distribution of conceptual and computational errors on the same subset of 100 questions. All categories have conceptual errors in most questions except for thermodynamics, transport phenomena, and fluid mechanics. Now, we will discuss some conceptual errors in different domains. The list of all questions subjected to analysis is provided in the GitHub repository of this work.

| Category | Total Questions | Conceptual error | | Computational error | |
|---|---|---|---|---|---|
| | | # questions | %age | # questions | %age |
| Thermodynamics | 11 | 4 | 36.36 | 8 | 72.73 |
| Atomic structure | 11 | 7 | 63.64 | 4 | 36.36 |
| Mechanical behavior | 11 | 7 | 63.64 | 4 | 36.36 |
| Material manufacturing | 11 | 8 | 72.73 | 3 | 27.27 |
| Electrical properties | 11 | 6 | 54.55 | 5 | 45.45 |
| Phase transition | 10 | 6 | 60.00 | 4 | 40.00 |
| Transport phenomena | 9 | 4 | 44.44 | 5 | 55.56 |
| Material Applications | 7 | 7 | 100.00 | 0 | 0 |
| Magnetic properties | 6 | 4 | 66.67 | 2 | 33.33 |
| Material characterization | 4 | 4 | 100.00 | 0 | 0 |
| Material processing | 4 | 4 | 100.00 | 0 | 0 |
| Miscellaneous | 3 | 3 | 100.00 | 0 | 0 |
| Fluid mechanics | 2 | 0 | 0 | 2 | 100.00 |

**Table 4.** Types of the error made by GPT-4-CoT on 100 questions classified according to domain expertise required to solve them

Fig. 4 (a) shows an example of the conceptual error made on a question related to thermodynamics. In this question, instead of considering the coefficient of thermal expansion the same in the planar dimension, it considered the coefficient of thermal expansion in the perpendicular direction as the same in one of the planar directions. Mathematically, instead of obtaining the final coefficient using 2 x parallel + perpendicular coefficients, GPT-4-CoT used parallel + 2 x perpendicular leading to an incorrect answer. While solving a question on atomic structure, as given in Fig. 8(b), GPT-4-CoT mistook the relation between lattice parameter (a) and atomic diameter (D) as $a = \frac{\sqrt{3}}{2}D$ instead of $a = \frac{2}{\sqrt{3}}D$. In a question on the electrical properties of materials (Fig. 4(c)), the GPT-4-CoT answered that all the given statements were correct. Hence, it could not choose from the four options given as answers. According to the materials science domain and the Wikipedia entry of Pourbaix diagrams, one of their major limitations is that these diagrams do not estimate actual corrosion rates; also, these diagrams cannot be used while studying corrosion due to chloride ions. Hence, the statement *R* is incorrect, making *(C)* the correct choice. While solving the question shown in Fig. 4(d), GPT-4-CoT did not convert the lattice parameter into the atomic diameter and considered them as same while using it in the formula required for solving the problem. For a question on materials manufacturing, GPT-4-CoT retrieved the functions of (P) blast furnace slag and (R) Torpedo car as opposite, thus leading to a wrong answer C when the correct option was A.

| | |
|---|---|
| (a) Thermodynamics | At 1000 K, the linear thermal expansion coefficients of graphite, parallel and perpendicular to the graphite layers, are 0.8 * $10^{-6} K^{-1}$ and 29 * $10^{-6} K^{-1}$, respectively. The percentage increase in the volume of graphite when heated from 900 K to 1100 K is? (round off to 2 decimal places) |
| (b) Atomic structure | Chromium has the bcc structure with atomic diameter of 2.494 Å. Find the first diffraction peak position ($2\vartheta$) for Cu K$\alpha$ radiation with a wavelength of 1.54 Å<br><br>(A) 21.76° (B) 33.05° (C) 44.43° (D) 66.10° |
| (c) Electrical properties | Identify the correct combination of the following statements:<br><br>P. Hydrogen electrode is a standard used to measure redox potentials<br>Q. Activation polarization refers to electrochemical processes controlled by reaction sequence at metal-solution interface<br>R. Potential-pH diagrams can be used to predict corrosion rates of metals<br>S. Cathodic protection can use sacrificial anodes such as magnesium<br><br>(A) P, Q and R (B) Q, R and S (C) P, Q and S (D) P, R and S |
| (d) Magnetic properties | Calculate the saturation magnetization (Msat) for bcc iron of lattice parameter 2.866 Å.<br><br>(A) 0.79 * $10^6$ A/m (B) 1.5 * $10^6$ A/m (C) 3.15 * $10^6$ A/m (D) 4.73 * $10^6$ A/m |
| (e) Material manufacturing | Match the reactors / refining sites in Column I with the corresponding refining processes in Column II.<br>Column I: [(P) Blast furnace runner, (Q) AOD, (R) Torpedo car, (S) BOF]<br>Column II: [1. De-carburization, 2. External De-sulfurization, 3. De-phosphorization, 4. External De-siliconization]<br><br>Options: (A) P-4, Q-1, R-2, S-3 (B) P-4, Q-2, R-3, S-1 (C) P-2, Q-1, R-4, S-3 (D) P-1, Q-3, R-2, S-4 |

**Figure 4.** Visualizing some of the questions where GPT-4-CoT made conceptual errors in the solution.

*Comparative analysis*

Finally, to answer the third question raised in this work, i.e., what factors limit the performance of LLMs on MaScQA, we visualize the mistakes made by GPT-3.5-CoT and the solution provided by GPT-4-CoT. Fig. 5 shows one example where GPT-4-CoT yielded the correct solution. If we check the Wikipedia page for phase rule (cite), the first expression is the one that is proposed as a solution by GPT-3.5. However, GPT-4 reaches the correct solution, also available on the same Wikipedia page. Although the dataset details on which these models are trained are unknown to the users, it is assumed that openly available sources like Wikipedia are a common dataset used by researchers while training such language models[18,30]. Thus, it is interesting to note that while GPT-3.5 depicts a shallow understanding of concepts, GPT-4 can provide a deeper understanding based on the context.

> At a constant atmospheric pressure, the number of phases, P which coexist in a chosen system at equilibrium, is related to the number of components, C in the system and the degree of freedom, F by (A) P+F=C-2 (B) P+F=C+2 (C) P+F=C+1 (D) P+F=C-1

> GPT-3.5-CoT: The Gibbs phase rule states that: F = C - P + 2 where F = degrees of freedom C = number of components P = number of phases Rearranging the equation: P + F = C + 2 Substituting the given values in this equation: P + F = C + 2 (at constant atmospheric pressure) Therefore, the answer is (B) P+F=C+2.

> GPT-4-CoT: The relationship between the number of phases (P), the number of components (C), and the degree of freedom (F) in a system at equilibrium is described by the Gibbs Phase Rule. The equation for the Gibbs Phase Rule is: P + F = C + 2 - Φ where Φ is the number of non-compositional variables (such as pressure, temperature, etc.) that can be changed independently without altering the system's phase equilibrium. When the problem states that the atmospheric pressure is constant, it means that pressure is not a variable that can be changed independently since it remains the same throughout the system. Consequently, Φ = 1 in this scenario. We are given that the atmospheric pressure is constant, so we can rewrite the equation as: P + F = C + 2 - 1 P + F = C + 1 Now we can see that the correct answer is (C) P + F = C + 1. Answer: [(C) P + F = C + 1]

**Figure 5.** Visualizing output of GPT models on a sample MCQ question.

The matching-type questions require understanding different topics and then the ability to interlink them. An example of a matching question with the solution as per GPT-3.5-CoT and

GPT-4-CoT is shown in Fig. 6. The scores in Table 1 indicated the exceptionally high performance of GPT-4 models in answering the matching-type question, which is more than two times the performance of GPT-3.5 models. It can be seen from the response of GPT-3.5-CoT that it is only able to determine the material properties required for the missile cone heads. Interestingly, GPT-3.5-CoT tries to arrive at the correct answer by eliminating the options. In contrast, GPT-4-CoT relied on understanding the topics and answering the question after inter-relating the previous information. This reinforces the idea that GPT-3.5 has a shallow understanding of the concepts.

> Match the composite in Column I with the most suitable application in Column II.
> Column I : [(P) Glass fibre reinforced plastic, (Q) SiC particle reinforced Al alloy, (R) Carbon-carbon composite, (S) Metal fibre reinforced rubber]
> Column II: [(1) Missile cone heads, (2) Commercial automobile chasis, (3) Airplane wheel tyres, (4) Car piston rings, (5) High performance skate boards]
>
> Options: (A) P-4, Q-5, R-1, S-2 (B) P-3, Q-5, R-2, S-4 (C) P-5, Q-4, R-1, S-3 (D) P-4, Q-2, R-3, S-1

> **GPT-3.5-CoT:** Composite | Application --- | --- P - Glass fibre reinforced plastic | 3 - Airplane wheel tyres Q - SiC particle reinforced Al alloy | 4 - Car piston rings R - Carbon-carbon composite | 1 - Missile cone heads S - Metal fibre reinforced rubber | 2 - Commercial automobile chasis
> Explanation: - Glass fiber reinforced plastic is known for its high strength-to-weight ratio and resistance to water and corrosion, making it suitable for airplane wheel tires (Option A and C eliminated).
> - SiC particle reinforced Al alloy is commonly used in high-performance pistons and piston rings because it has low thermal expansion, high thermal conductivity, and is resistant to wear and abrasion, making it suitable for car piston rings (Option C eliminated).
> - Carbon-carbon composites are known for their high strength and heat resistance, making them suitable for missile cone heads, which must withstand high temperatures and stresses during operation (Option B eliminated).
> - Metal fiber reinforced rubber is commonly used in commercial automobile chassis to improve the mechanical properties of rubber and is used to reduce noise and vibration (Option D eliminated). Therefore, the correct answer is (A) P-3, Q-5, R-1, S-2.

> **GPT-4-CoT:** To correctly match the composite materials in Column I with their most suitable applications in Column II, we need to understand the properties of each composite material and the requirements for each application.
> Column I: (P) Glass fibre reinforced plastic (GFRP) - Lightweight, strong, and resistant to corrosion.
> (Q) SiC particle reinforced Al alloy - High strength, stiffness, wear resistance, and thermal stability.
> (R) Carbon-carbon composite - High-temperature resistance, lightweight, and high strength.
> (S) Metal fibre reinforced rubber - Improved tensile strength, abrasion resistance, and flexibility.
>
> Column II: (1) Missile cone heads - Require lightweight, heat-resistant, and high strength materials.
> (2) Commercial automobile chassis - Require high-strength, lightweight, and corrosion-resistant materials.
> (3) Airplane wheel tyres - Require materials with good tensile strength, abrasion resistance, and flexibility.
> (4) Car piston rings - Need high strength, wear resistance, and thermal stability.
> (5) High-performance skateboards - Require lightweight, strong, and corrosion-resistant materials.
>
> Now let's try to match the materials with the applications: (P) GFRP - Its lightweight, strong, and corrosion-resistant properties make it suitable for high-performance skateboards. So, P matches with 5. (Q) SiC particle reinforced Al alloy - High strength, wear resistance, and thermal stability make this material suitable for car piston rings. So, Q matches with 4. (R) Carbon-carbon composite - High-temperature resistance, lightweight, and high strength properties make this material well-suited for missile cone heads. So, R matches with 1. (S) Metal fibre reinforced rubber - Improved tensile strength, abrasion resistance, and flexibility make this material suitable for airplane wheel tyres. So, S matches with 3. Thus, the correct matching is P-5, Q-4, R-1, S-3, which corresponds to option (C).

**Figure 6.** Visualizing output of GPT models on a sample matching type question

An example of a numerical question with multiple options is shown in Fig. 7. The GPT-3.5-CoT solution used the correct concept but made calculation errors leading to a final incorrect answer. However, GPT-4-CoT used the correct concept and did not make calculation mistakes. It is observed in Table 3 that both GPT-4 and GPT-4-CoT achieve similar accuracy in answering MCQN questions. The red-colored text in the GPT-3.5-CoT solution shows the source of the error, which led to an incorrect answer.

> A 20 kN tensile load is applied axially to a steel bar of cross-sectional area 8 $cm^2$ and 1 $m$ length. The Young's modulus of steel ($E_{(steel)}$) is 200 GPa, and of aluminium ($E_{(Al)}$) is 70 GPa. The Poisson's ratio (ν) can be taken as 0.3. When the same load is applied to an aluminium bar, it is found to give same elastic strain as the steel. Calculate the cross-sectional area of the aluminium bar.
> (A) 11.43 $cm^2$ (B) 14.93 $cm^2$ (C) 18.26 $cm^2$ (D) 22.86 $cm^2$

> **GPT-3.5-CoT:** Given data, Force, F = 20 kN = 20 x 10³ N, Length, L = 1 m, Cross-sectional area of steel bar, $A_{\{steel\}}$ = 8 $cm^2$, Young's modulus of steel, $E_{steel}$ = 200 $GPa$ = 200×10⁹ $N/m^2$, Young's modulus of aluminium, $E_{\{Al\}}$ = 70 $GPa$ = 70 ×10⁹ $N/m^2$, Poisson's ratio, $\nu$ = 0.3
> Let $\epsilon_{steel}$ and $\epsilon_{Al}$ be the elastic strain in the steel and aluminium bars respectively. We also know that strain, $\Delta L/L = \epsilon_L/1 = \epsilon$ Where, $\Delta L$ is change in length and L is original length. Now, we can first calculate the stress in the steel bar as, $\sigma_{steel} = F/A_{steel}$ And then calculate the elastic strain in the steel bar as, $\epsilon_{steel} = \sigma_{steel}/E_{steel}$
> Using the fact that elastic strain is same for aluminium and steel, we can calculate the force that needs to be applied for the aluminium bar as follows, $\epsilon_{Al} = \frac{\sigma_{Al}}{E_{Al}}$, therefore, $\sigma_{Al} = \epsilon_{Al} \times E_{Al}$
> because $\epsilon_{Al} = \epsilon_{steel}$, therefore, $\sigma_{Al} = \epsilon_{steel} \times E_{Al}$ therefore $F_{Al} = \sigma_{Al} \times A_{Al}$
> because $F_{Al} = F = 20 \times 10^3$ N therefore $\sigma_{Al} \times A_{Al} = 20 \times 10^3$ N
> Putting the value of $\sigma_{Al}$ from above, we get $A_{Al} = \frac{20 \times 103 \text{ N}}{\epsilon_{steel} \times E_{Al}}$
> Now, we need to calculate the elastic strain in the steel bar. This can be done as follows, $\sigma_{steel} = F/A_{steel}$
> $\epsilon_{steel} = \sigma_{steel}/E_{steel}$, therefore $\epsilon_{steel} = F/A_{steel} \times E_{steel}$, therefore $\epsilon_{steel}$ = 20 x 10³ N/8 x 10⁻⁴ m² x 200 x 10⁹ $N/m^2$
> therefore $\epsilon_{steel} = 1.25 \times 10^{-4}$
> Now, we can put the value of $\epsilon_{steel}$ in our equation for $A_{Al}$ that we derived earlier to get the final answer.
> $A_{Al} = \frac{20 \times 103 \text{ N}}{\epsilon_{steel} \times E_{Al}} = \frac{\mathbf{20 \times 103 \text{ N}}}{\mathbf{1.25 \times 10^{-4} \times 70 \times 10^9 \text{ N/m}^2}}$
> $A_{Al}$ = 14.93 cm² (approx).
> Therefore, the cross-sectional area of the aluminium bar is approximately 14.93 cm². Hence, the correct answer is option (B)14.93 cm² .

> **GPT-4-CoT:** To solve this problem, we first need to find the stress and strain in the steel bar when the load is applied. Then, we'll find the stress in the aluminium bar using the same strain, and finally, we'll calculate the cross-sectional area of the aluminium bar.
> Step 1: Find the stress in the steel bar Stress in the steel bar ($\sigma_{(steel)}$) is given by the formula: $\sigma_{(steel)} = \frac{Load}{Area}$
> Given load = 20 kN = 20,000 N, and the cross-sectional area of steel bar = 8 $cm^2$ = 8×10⁻⁴$m^2$. So, $\sigma_{steel} = \frac{20,000}{8 \times 10^{-4}} = 2.5 \times 10^7 Pa$
> Step 2: Find the strain in the steel bar Strain in the steel bar ($\varepsilon_{(steel)}$) is given by the formula: $\varepsilon_{(steel)} = \frac{\sigma_{steel}}{E_{(steel)}}$ Given $E_{(steel)}$ = 200 GPa = 200×10⁹$Pa$, So, $\varepsilon_{(steel)} = \frac{2.5 \times 10^7}{200 \times 10^9} = 1.25 \times 10^{-4}$
> Step 3: Find the stress in the aluminium bar Since the strain in the aluminium bar is the same as the strain in the steel bar: $\varepsilon_{(Al)} = \varepsilon_{(steel)}$
> Stress in the aluminium bar ($\sigma_{(Al)}$) is given by the formula: $\sigma_{Al} = E_{Al} \times \varepsilon_{(Al)}$
> Given $E_{(Al)}$ = 70 GPa = 70×10⁹$Pa$, and $\varepsilon_{(Al)} = 1.25 \times 10^{-4}$, So, $\sigma_{(Al)} = 70 \times 10^9 \times 1.25 \times 10^{-4} = 8.75 \times 10^6$ Pa
> Step 4: Calculate the cross-sectional area of the aluminium bar Let the cross-sectional area of the aluminium bar be A. Stress in the aluminium bar is given by the formula: $\sigma_{Al} = \frac{Load}{Area}$.
> From the above equation, we get: Area (A) = $\frac{Load}{\sigma_{Al}}$ Given load = 20,000 N, and $\sigma_{Al} = 8.75 \times 10^6 Pa$, So, A = $\frac{20,000}{8.75 \times 10^6}$ = 2.2857×10⁻³ $m^2$. Converting to $cm^2$, we get: A = 2.2857×10⁻³× 10⁴ = 22.857 $cm^2$. Therefore, the cross-sectional area of the aluminium bar is approximately 22.86 $cm^2$. The correct option is (D) 22.86 $cm^2$.

**Figure 7**. Visualizing the output of GPT models on a numerical question with multiple options

Now, we show the comparison of the solution by GPT-3.5-CoT and GPT-4-CoT on a sample numerical question (NUM) in Fig. 8 related to platinum's crystal structure. Both models applied the correct concept. However, GPT-3.5-CoT made a calculation mistake in obtaining the interplanar distance *"d"*, which is highlighted in boldface and red color in Fig. 8. Calculation mistakes are a known issue with such kinds of LLMs from the literature[7–9,18,31] where similar order of accuracy was achieved on numerical questions solving tasks. The low accuracy of LLMs may also imply a lack of material science concepts previously observed in MCQ and MATCH-type questions in addition to calculation in capability.

---

The expected diffraction angle (in degrees) for the first order reflection from the (113) set of planes for face centered cubic Pt (lattice parameter = 0.392 nm) using monochromatic radiation of wavelength 0.1542 nm is

---

**GPT-3.5-CoT**: To solve this problem, we will use Bragg's law: $n\lambda = 2d\sin\vartheta$ where n is the order of diffraction (in this case, n = 1), λ is the wavelength of the radiation (0.1542 nm), d is the interplanar spacing, and ϑ is the diffraction angle we are trying to find. To find d for the (113) planes of face-centered cubic Pt, we can use the formula: $d = a / \sqrt{h^2 + k^2 + l^2}$ where a is the lattice parameter (0.392 nm), and h, k, and l are the Miller indices for the set of planes we are interested in (113): **d = 0.392 nm / √(1^2 + 1^2 + 3^2) = 0.136 nm** Now we can use Bragg's law to find ϑ: sinϑ = nλ / 2d = (1)(0.1542 nm) / (2)(0.136 nm) = 0.566 ϑ = sin⁻¹(0.566) = 34.6° Therefore, the expected diffraction angle for the first order reflection from the (113) set of planes for face centered cubic Pt using monochromatic radiation of wavelength 0.1542 nm is 34.6°. Answer: [34.6°]

---

**GPT-4-CoT**: To solve this problem, we'll use Bragg's Law, which relates the angle of reflection to the interplanar distance (d-spacing) and the wavelength of the radiation. Bragg's Law is given by: $n\lambda = 2d \ast \sin\vartheta$ where: - n is the order of reflection (in this case, n = 1 for first order) - λ is the wavelength of the radiation (in this case, λ = 0.1542 nm) - d is the interplanar distance - ϑ is the angle of reflection (the value we want to calculate) First, we need to find the interplanar distance (d-spacing) for the (113) set of planes in a face-centered cubic (FCC) lattice. To do this, we'll use the formula for d-spacing in a cubic lattice: $d = (a) / \sqrt{h^2 + k^2 + l^2}$ where: - a is the lattice parameter (in this case, a = 0.392 nm) - h, k, and l are the Miller indices for the set of planes (in this case, h = 1, k = 1, and l = 3) Plug in the values into the d-spacing formula: d = (0.392 nm) / sqrt((1)^2 + (1)^2 + (3)^2) d = (0.392 nm) / sqrt(1 + 1 + 9) d = (0.392 nm) / sqrt(11) d ≈ 0.1181 nm Now that we have the d-spacing, we can use Bragg's Law to find ϑ: 1 * (0.1542 nm) = 2 * (0.1181 nm) * sinϑ Rearrange the equation to solve for sinϑ: sinϑ = (0.1542 nm) / (2 * 0.1181 nm) sinϑ ≈ 0.6531 Now, to find the angle ϑ, we need to take the inverse sine (arcsin) of the sinϑ value: ϑ = arcsin(0.6531) ϑ ≈ 40.687° So, the expected diffraction angle for the first order reflection from the (113) set of planes in a face centered cubic Pt lattice using monochromatic radiation of wavelength 0.1542 nm is approximately 40.687°. Answer: [40.687]

**Figure 8**: Visualizing output of GPT models on a sample numerical question

Now, we will discuss the performance of GPT-4-CoT from the materials science domain perspective. The topics in Table 2 are arranged in decreasing order of the total number of questions in each category. The maximum percentage of incorrect questions is in questions under the *electrical* topic. The incorrectly answered questions require solving questions related to battery cells, the redox reactions, or identifying the potentials between the electrodes. The number of numerical questions answered wrong is 3 – 5 times of the other type of questions. Regarding questions related to the mechanical behavior of materials, GPT-4-CoT has the second-worst performance. Out of 53 incorrectly answered questions, 34 are numerical questions. The questions where mistakes happened were based on concepts of the materials' stress-strain curve, fracture mechanics, and creep behavior. The *thermodynamics* category has a maximum number of questions and quite a high percentage of incorrectly answered questions (~46%). The incorrect questions require understanding concepts of formation energy, specific heat, heat transfer, and chemical equations, to name a few, and solving complex equations correctly. The category of *atomic structure* has ~42% incorrectly answered, mostly related to questions on the analysis of X-Ray diffraction studies to identify the crystal structure of the materials. This reflects that LLMs are unable to correlate theoretical concepts with experimental outcomes. The category *magnetism* has fewer questions (15), of which only eight are correct. The performance of LLMs in answering these questions reflects their inability to retrieve related concepts like magnetic moment and saturation magnetizations and avoid numerical errors. In *phase transitions,* the incorrectly answered questions (~41%) are related to solving for the composition of different phases after the transitions and conditions required for phase transition. The next category is *transport,* where the incorrectly answered questions (~38%) required understanding diffusion phenomena and concepts of thermodynamics and battery cell reactions.

To summarise, the CoT prompting cannot significantly improve the LLM performance as the mistakes are mainly conceptual. This makes a strong case for a domain-specific LLM for materials and potentially domain-specific alternate prompting strategies. Further, for questions where the LLMs give the incorrect response due to computational error, the solution involved unit conversions, logarithms, and exponentials and had numbers with multiplying factors (e.g., $10^{10}$). There have been recent works in the literature that suggest methods for improving calculations and for improving on concept-based mistakes[33]. Introducing such heuristics while prompting can help researchers in two ways: (1) probe the existing LLMs more deeply, (2) generate datasets to train LLMs with lesser parameters, thus, making the use of these models economical. Hence, this answers the third research question (limiting factors for LLMs) raised in this work.

**Conclusion**

Due to the increasing availability of large datasets and computation capabilities, developing an LLM is becoming relatively easier. In materials discovery, machine learning and natural language processing have played an instrumental role in identifying new materials or existing materials for a new application, discovering an optimal synthesis pathway, and planning. We are living in an era where machine learning, humans, and machines are working together in the pipeline of discovering new materials. At this juncture, it is crucial to ask how well LLMs understand the materials science domain, as the answer to this will determine their applications in such pipelines. To this end, our new dataset, MaScQA, used to test the mental abilities required to solve the questions and understand the materials science domain and their interrelated concepts, will provide a means to gain deeper insights. We observed that the LLMs make both numerical and conceptual mistakes. There are several core materials science domains where LLMs show poor performance, such as the atomic and crystal structure of materials and their electrical, magnetic, and thermodynamic behavior. Hence, to enable their use in the materials discovery pipeline, the language models must be finetuned on a domain-specific dataset.

Moreover, the performance of the LLMs on MaScQA can enable a deeper understanding of the lacunae in the LLMs, thereby providing new research avenues. For instance, LLMs' poor performance in NUM questions suggests that a pipeline connecting the LLM to a math calculator can potentially yield improved results. Further, the conceptual mistakes made by the LLMs suggest areas where further improvements are required. The materials science domain is a field that derives concepts from physics, chemistry, and mechanics. Therefore, a benchmark like MaScQA will allow the researchers to benchmark their results against a standard dataset. Further, the correct solutions can help researchers create a new dataset for training lightweight models, which are economical and hence, can be easily deployed on low-memory industrial devices for materials discovery and their usage for educational purposes.